\documentclass[letterpaper]{article} 
\usepackage{aaai2027}  
\usepackage[hyphens]{url}  
\usepackage{graphicx} 
\usepackage{natbib}  
\usepackage{caption} 
\usepackage{algorithm}
\usepackage{algorithmic}

\usepackage{newfloat}
\usepackage{listings}
\DeclareCaptionStyle{ruled}{labelfont=normalfont,labelsep=colon,strut=off} 
\floatstyle{ruled}
\newfloat{listing}{tb}{lst}{}
\floatname{listing}{Listing}

\usepackage{booktabs}
\usepackage{multirow}
\usepackage{amsmath,amsfonts,amssymb,amsthm}
\nocopyright
\title{Noise-Aware Shrinkage for Differentially Private \\Zeroth-Order Fine-Tuning of Large Language Models}
\author{
    Lele Zheng\textsuperscript{},
    Weifeng Kong\textsuperscript{},
    Xinyi Zhang\textsuperscript{},
    Ke Cheng\textsuperscript{},
    Tao Zhang\textsuperscript{},
    Yulong Shen\textsuperscript{}
}
\affiliations{
    School of Computer Science and Technology, Xidian University\\
}

\begin{document}

\maketitle

\begin{abstract}

Differentially private zeroth-order optimization (DP-ZO) enables memory-efficient private fine-tuning of large language models using only forward evaluations. Existing aggregation-based DP-ZO methods reconstruct model updates at a fixed scale, ignoring that the strength of useful signals varies throughout training. Consequently, noise-dominated updates may receive excessive weight and degrade model utility. To address this issue, we propose SAGE, a noise-aware shrinkage method that adaptively attenuates privatized estimates according to their estimated signal quality. SAGE subtracts the known Gaussian noise variance from the observed second moment to estimate the underlying signal energy, stabilizes this estimate through temporal tracking, and compares its current signal-to-noise level with a warm-up reference to derive a bounded shrinkage factor. As pure post-processing, SAGE requires neither additional privacy budget nor model queries and introduces only constant additional state. Our theoretical analysis shows that shrinkage reduces the quadratic update-risk term faster than the linear descent term, preserving useful descent while limiting the influence of noise-dominated updates. Experiments on RoBERTa-large, OPT-1.3B, and OPT-6.7B demonstrate that SAGE outperforms existing baselines in most settings under the same privacy budgets while preserving the forward-only memory efficiency of DP-ZO.
\end{abstract}

\section{Introduction}

Fine-tuning large language models (LLMs) for specific tasks and domains has become a common practice in modern machine learning. Traditional fine-tuning relies on backpropagation and first-order optimizers, incurring substantial memory overhead from activations, gradients, and optimizer states~\citep{kingma2017adammethodstochasticoptimization,loshchilov2018decoupled,JMLR:v12:duchi11a}. To alleviate this bottleneck, MeZO introduces zeroth-order optimization (ZO) into LLM fine-tuning~\citep{NEURIPS2023_a6278101}. By estimating update directions through only two forward evaluations~\citep{119632} and regenerating random perturbations from their seeds, MeZO avoids backpropagation and reduces the training memory footprint to a level close to inference.
When fine-tuning involves sensitive data, formal privacy protection is also essential. Building on this forward-only paradigm, differentially private zeroth-order optimization (DP-ZO) clips and perturbs the data-dependent directional response, enabling memory-efficient and privacy-preserving fine-tuning of LLMs~\citep{tang2025private}.

A central challenge in DP-ZO is the distortion introduced by clipping. Gradient scalar estimates can vary substantially across examples and random directions, making it difficult to select an appropriate fixed clipping threshold. A small threshold may excessively truncate informative signals, whereas a large threshold increases the sensitivity bound and consequently requires stronger privacy noise. DP-AggZO mitigates this issue by jointly clipping a \(K\)-dimensional vector of directional estimates before Gaussian perturbation \citep{309502}. As \(K\) increases, the aggregate norm becomes more stable, reducing clipping error while preserving the forward-only and memory-efficient structure of DP-ZO.

Despite these advances, existing aggregation-based DP-ZO methods directly use the privatized gradient-scalar estimate vector obtained from multiple random directions, implicitly applying the same nominal scale across training steps. However, the energy of the underlying pre-noise aggregate can vary substantially during optimization
~\citep{pmlr-v139-asi21a,NEURIPS2024_49c466cc,NEURIPS2025_511f4fcf},
whereas the Gaussian noise variance is determined by the privacy mechanism rather than the current signal magnitude. Consequently, the effective signal-to-noise ratio of the released vector can be highly non-stationary. When the signal becomes weak, privacy noise
accounts for a larger fraction of the update, increasing
update risk and degrading optimization.

\begin{figure}[t]
    \centering
    \includegraphics[width=\columnwidth]{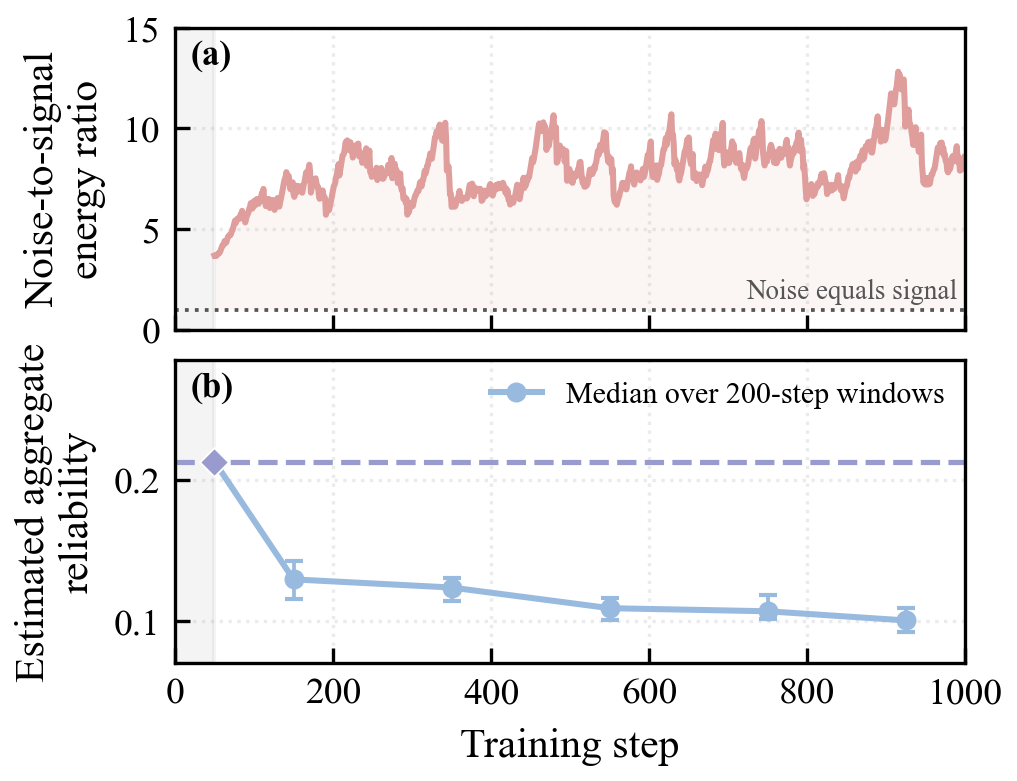}
    \caption{
Non-stationary reliability of privatized gradient scalar estimates in DP-ZO
fine-tuning. The signal quality varies throughout training under a fixed
Gaussian noise level, while existing methods apply the same update scale
regardless of estimate reliability.
}
    \label{fig:nonstationary_reliability}
\end{figure}

To examine this non-stationarity, we analyze the privatized
gradient scalar estimates from a representative SNLI
fine-tuning run. As shown in
Figure~\ref{fig:nonstationary_reliability}, Gaussian privacy
noise dominates the signal, with a median noise-to-signal
energy ratio of \(7.8\). Here, reliability measures the
estimated signal strength relative to the Gaussian privacy
noise. It decreases from \(0.213\) at the end of warmup to
approximately \(0.100\) in the late stage. These results show
that signal quality varies during training, while fixed-scale
updates use the same nominal scale across iterations.
This mismatch motivates adaptive shrinkage. Scaling the
privatized estimate vector by \(m_t\in(0,1]\) scales the
useful descent term by \(m_t\), but the quadratic update-risk
term by \(m_t^2\). Thus, attenuating low-reliability
estimates can suppress noise-induced risk more strongly than
useful descent.

Based on the above observations, we propose \textbf{SAGE}, a noise-aware shrinkage method for
adaptive update scaling in aggregation-based DP-ZO. Given a
released privatized gradient scalar estimate vector
\(\tilde{s}_t=u_t+\xi_t\), its observed energy
\(\|\tilde{s}_t\|_2^2/K\) contains an additive noise floor
\(\tau_t^2\) determined by the Gaussian mechanism
~\citep{10.1145/3219819.3220076,pmlr-v80-balle18a,tang2023dpadam,tang2023dpadambcdpadamactuallydpsgd}.
SAGE removes this known noise floor, tracks the corrected
signal energy with a warmup-anchored exponential moving
average, and computes a shrinkage factor
\(m_t\in(0,1]\) to rescale the released estimate as
\(\bar{s}_t=m_t\tilde{s}_t\). SAGE never amplifies the
estimate and, as pure post-processing, requires no additional
queries or privacy cost. Its lightweight design can be seamlessly integrated into existing aggregation-based DP-ZO methods without modifying their optimization pipeline or privacy accounting.We apply SAGE to the privatized gradient scalar estimate vector and evaluate it on RoBERTa-large and OPT models
across multiple text classification and natural language
understanding tasks. The results show that SAGE improves
private fine-tuning utility under identical privacy budgets
and training settings.


Our contributions are summarized as follows:
\begin{itemize}
    \item We identify and characterize the non-stationary
    reliability of Gaussian-privatized gradient scalar
    estimates, showing that fixed-scale updates do not account
    for changes in signal quality during training.
    \item We propose SAGE, a noise-aware shrinkage method
    that corrects the observed energy for the known Gaussian
    noise floor and adaptively scales privatized estimates
    through pure post-processing.
    \item We show theoretically that shrinkage suppresses the quadratic update-risk term more rapidly than the linear descent term, and empirically validate SAGE across RoBERTa-large and OPT models under the same privacy budgets and training settings.
\end{itemize}

\section{Related Work}
\label{sec:related}

\paragraph{First-Order Differentially Private Fine-Tuning.}
First-order optimizers such as Adam, AdamW and AdaGrad
~\citep{kingma2017adammethodstochasticoptimization,loshchilov2018decoupled,JMLR:v12:duchi11a} are widely used for model training and fine-tuning.
To protect sensitive training data, DP-SGD clips per-example gradients and injects Gaussian noise~\citep{10.1145/2976749.2978318}, while DP-Adam and
DP-AdamW provide differentially private variants of Adam and
AdamW, respectively~\citep{tang2023dpadam,sun2025dpadamw}.
Despite their strong utility, these methods require backpropagation,
per-example gradients, and parameter-dimensional optimizer states,
making their memory cost increasingly restrictive with growing LLM sizes.

\paragraph{Differentially Private Zeroth-Order Optimization.}
Following the SPSA framework~\citep{119632}, MeZO~\citep{NEURIPS2023_a6278101} reduces the memory overhead of LLM fine-tuning by replacing
backpropagation with zeroth-order gradient estimates computed using only forward passes. 
For private fine-tuning, DPZero~\citep{pmlr-v235-zhang24af} and
DP-ZO~\citep{tang2025private} privatize scalar ZO responses, while
DP-AggZO~\citep{309502} aggregates multiple independent ZO
estimates to reduce the clipping error in DP-ZO and improve the accuracy--privacy tradeoff. However, existing methods still use privatized gradient scalar estimates at a fixed scale, so noise-dominated updates may receive excessive weight and degrade model utility.

\paragraph{Noise-Floor Correction under Differential Privacy.}
A related line corrects DP-induced bias in adaptive optimizers.
DP-Adam~\citep{tang2023dpadam} and DP-AdamBC~\citep{tang2023dpadambcdpadamactuallydpsgd} correct the
second-moment estimator by subtracting the known Gaussian noise
variance, while related noise-bias correction
strategies are further explored in
DP-FedAdamW~\citep{Liu_2026_CVPR} and
FIBER~\citep{dm2026fiber}. These methods correct per-coordinate
first-order preconditioners, usually with $O(d)$ optimizer state.

\section{Preliminaries}
\label{sec:preliminaries}

\paragraph{Differential privacy.}
Differential privacy (DP) provides a formal framework for limiting the privacy leakage of individual records. Two datasets \(D\) and \(D'\) are neighboring if one can be obtained from the other by adding or removing a single record. A randomized mechanism \(\mathcal{M}\) satisfies \((\varepsilon,\delta)\)-differential privacy if, for all neighboring \(D,D'\) and any measurable output set \(\mathcal{O}\),
\begin{equation}
\Pr[\mathcal{M}(D)\in\mathcal{O}]
\le
e^{\varepsilon}\Pr[\mathcal{M}(D')\in\mathcal{O}]
+ \delta .
\end{equation}

\paragraph{Zeroth-order and differentially private zeroth-order optimization.}
Zeroth-order (ZO) optimization estimates update directions using only function values~\citep{9186148}. At step \(t\), ZO samples a random perturbation direction \(z_t\sim\mathcal{N}(0,I_d)\) and evaluates the loss at \(\theta_t+\mu z_t\) and \(\theta_t-\mu z_t\). For an example \(x\), the resulting gradient scalar estimate is
\begin{equation}
\Delta_{z_t}(\theta_t;x)
=
\frac{
\ell(\theta_t+\mu z_t;x)
-
\ell(\theta_t-\mu z_t;x)
}{2\mu},
\end{equation}
where \(\theta_t\) denotes the model parameters, \(\mu\) is
the perturbation scale, and \(\ell(\theta;x)\) is the loss on
example \(x\).

Differentially private zeroth-order optimization (DP-ZO) clips and perturbs this data-dependent gradient scalar estimate before forming updates. 

\paragraph{Differentially private aggregated zeroth-order optimization.}
Differentially Private Aggregated Zeroth-Order Optimization
(DP-AggZO) extends DP-ZO by sampling \(K\) independent Gaussian
directions \(z_{t,1},\ldots,z_{t,K}\) and computing one gradient
scalar estimate along each direction. Let
\(\Delta_{t,k}(x)=\Delta_{z_{t,k}}(\theta_t;x)\). For each example \(x\), these
estimates form the \(K\)-dimensional gradient scalar estimate vector
\begin{equation}
    \Delta_{\mathrm{agg}}(\theta_t;x)
    =
    \frac{1}{K}
    \left(
        \Delta_{t,1}(x),
        \ldots,
        \Delta_{t,K}(x)
    \right).
\end{equation}
Given a clipping threshold \(C\), DP-AggZO jointly clips this vector as
\begin{equation}
    \widehat{\Delta}_{\mathrm{agg}}(\theta_t;x)
    =
    \min
    \left(
        1,
        \frac{C}
        {
            \|
            \Delta_{\mathrm{agg}}(\theta_t;x)
            \|_2
        }
    \right)
    \Delta_{\mathrm{agg}}(\theta_t;x).
\end{equation}
The clipped vectors are then aggregated over the sampled batch and
perturbed with Gaussian noise to obtain the privatized gradient scalar
estimate vector \(\tilde{s}_t\).

\section{Method}
\label{sec:method}

\begin{figure*}[t]
    \centering
    \includegraphics[width=\textwidth,height=0.30\textheight,keepaspectratio]{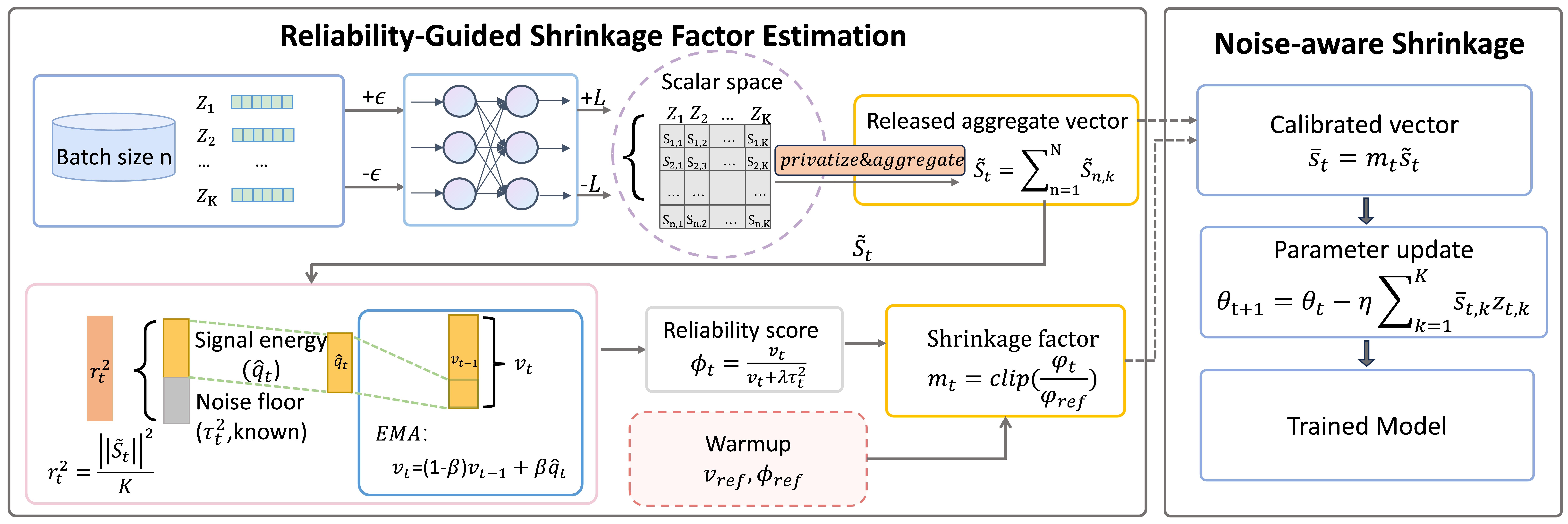}
    \caption{
    Overview of the SAGE framework. Starting from the released
privatized gradient-scalar vector, SAGE explicitly corrects
its observed energy for the known Gaussian noise floor,
temporally smooths the resulting signal estimate, and
converts it into a bounded, noise-aware shrinkage factor.
This factor adaptively rescales the released vector before
reconstructing the zeroth-order parameter update, effectively
suppressing unreliable, noise-dominated updates without
introducing additional model queries or privacy cost. The resulting update remains
forward-only and memory-efficient.
    }
    \label{fig:sage_overview}
\end{figure*}

This section presents SAGE, a noise-aware shrinkage method
for adaptive update scaling of privatized gradient scalar
estimates. As illustrated in Figure~\ref{fig:sage_overview}, SAGE operates entirely on the released
low-dimensional estimate vector.

Following DP-AggZO~\citep{309502}, at step \(t\) we obtain
a privatized gradient scalar estimate vector
\begin{equation}
    \tilde{s}_t=u_t+\xi_t,
    \qquad
    \xi_t\sim\mathcal{N}(0,\tau_t^2I_K),
\end{equation}
where \(u_t\in\mathbb{R}^K\) contains the clipped and
aggregated gradient scalar estimates before noise addition,
and \(\tau_t^2\) is the known per-coordinate Gaussian noise
variance. SAGE only post-processes \(\tilde{s}_t\), leaving
the queries, clipping rule, Gaussian mechanism, and privacy
accountant unchanged.

Alg.~\ref{alg:sage} summarizes the resulting procedure.
SAGE adds only noise-floor correction, scalar temporal tracking,
and bounded shrinkage to the released estimate vector.

\begin{algorithm}[tb]
\caption{SAGE: Noise-Aware Shrinkage of Gradient Estimates}
\label{alg:sage}
\small
\textbf{Input}: Dataset \(D\), steps \(T\), sampling rate \(q\), learning rate \(\eta\), perturbation scale \(\mu\), noise multiplier \(\sigma\), clipping threshold \(C\), directions \(K\), warmup \(W\), EMA rate \(\beta\), parameter \(\lambda\), floor \(\rho\), lower bound \(m_{\min}\), Laplace scale \(\psi\).
\begin{algorithmic}[1]
\STATE Release
\(
    \tilde n
    \leftarrow
    \max
    \left\{
        1,
        |D|+\operatorname{Lap}(0,\psi)
    \right\},
\)
and set
\(
    \tilde m
    \leftarrow
    q\tilde n.
\)
\STATE Initialize buffers \(\mathcal B_q,\mathcal B_\tau\leftarrow\emptyset\).
\FOR{\(t=1,\ldots,T\)}
    \STATE Sample batch \(B_t\) by Poisson sampling.
    \STATE Form and privatize the \(K\)-dimensional gradient scalar estimate vector:
    \[
    \tilde{s}_t
    \leftarrow
    \frac{1}{\tilde m}
    \left(
        \sum_{x\in B_t}
        \widehat{\Delta}_{\mathrm{agg}}(\theta_t;x)
        +
        \mathcal N(0,\sigma^2C^2I_K)
    \right).
    \]
    \STATE Set \(\tau_t^2\leftarrow(\sigma C/\tilde m)^2\).
    \STATE Estimate debiased signal energy:
    \[
        \widehat q_t
        \leftarrow
        \max\left(\|\tilde{s}_t\|_2^2/K-\tau_t^2,\rho\right).
    \]
    \IF{\(t\le W\)}
        \STATE Append \(\widehat q_t\) to \(\mathcal B_q\), append \(\tau_t^2\) to \(\mathcal B_\tau\).
        \STATE Set \(m_t\leftarrow 1\).
        \IF{\(t=W\)}
            \STATE \(v_{\rm ref}\leftarrow{\rm mean}(\mathcal B_q)\), \(\tau_{\rm ref}^2\leftarrow{\rm mean}(\mathcal B_\tau)\).
            \STATE \(v\leftarrow v_{\rm ref}\), \(\phi_{\rm ref}\leftarrow v_{\rm ref}/(v_{\rm ref}+\lambda\tau_{\rm ref}^2)\).
        \ENDIF
    \ELSE
        \STATE \(v\leftarrow(1-\beta)v+\beta\widehat q_t\).
        \STATE \(\phi_t\leftarrow v/(v+\lambda\tau_t^2)\).
        \STATE \(m_t\leftarrow{\rm clip}(\phi_t/\phi_{\rm ref},\,m_{\min},\,1)\).
    \ENDIF
    \STATE Apply shrinkage: \(\bar{s}_t\leftarrow m_t\tilde{s}_t\).
    \STATE Update:
    \[
        \theta_{t+1}
        \leftarrow
        \theta_t
        -
        \eta
        \sum_{k=1}^{K}
        \bar{s}_{t,k}z_{t,k}.
    \]
\ENDFOR
\RETURN Model parameters \(\theta\).
\end{algorithmic}
\end{algorithm}

\subsection{Estimating Reliability from Privatized Gradient Scalar Estimates}
\label{sec:reliability}

The key observation behind SAGE is that the reliability of the
released aggregate is not stationary during training. Although the
Gaussian noise variance is fixed by the privacy mechanism, the
magnitude of the underlying optimization signal changes as optimization
progresses. This motivates estimating the signal strength of the
privatized gradient scalar estimates and adapting the update scale
accordingly.

To estimate this reliability, SAGE exploits the known Gaussian noise
structure of the released estimates. We first compute the squared magnitude of the released estimate vector:
\begin{equation}
    r_t^2
    =
    \frac{\|\tilde{s}_t\|_2^2}{K}.
\end{equation}

The squared norm is used because the Gaussian perturbation introduces
a predictable additive component into the observed second moment. Given
\begin{equation}
    \tilde{s}_t=u_t+\xi_t,
    \qquad
    \xi_t\sim\mathcal{N}(0,\tau_t^2I_K),
\end{equation}
the conditional expectation satisfies
\begin{equation}
\begin{aligned}
    \mathbb{E}_{\xi}
    \left[
    r_t^2\mid u_t
    \right]
    &=
    \frac{1}{K}
    \mathbb{E}_{\xi}
    \left[
    \|u_t+\xi_t\|_2^2
    \mid u_t
    \right] \\
    &=
    \frac{\|u_t\|_2^2}{K}
    +
    \tau_t^2 .
\end{aligned}
\end{equation}

Therefore, the Gaussian mechanism contributes a known noise floor
\(\tau_t^2\) to the observed estimate energy. SAGE removes
this contribution and obtains a noise-floor-corrected estimate of the signal energy:
\begin{equation}
    \widehat q_t
    =
    \max
    \left(
    r_t^2-\tau_t^2,
    \rho
    \right),
\end{equation}
where \(\rho>0\) prevents degenerate estimates near the
noise level.

Since \(\widehat q_t\) is computed from a single release, it can fluctuate substantially across iterations. SAGE therefore tracks the temporal evolution of the estimated signal
energy using an exponential moving average:
\begin{equation}
    v_t
    =
    (1-\beta)v_{t-1}
    +
    \beta\widehat q_t .
\end{equation}

This moving average only tracks the temporal variation of the
reliability-related statistic and does not accumulate optimization
directions. Therefore, it introduces only a scalar state and preserves the memory efficiency of DP-ZO.

During the first \(W\) steps, SAGE applies no shrinkage and
collects:
\begin{equation}
    v_{\mathrm{ref}}
    =
    \frac{1}{W}
    \sum_{j=1}^{W}
    \widehat q_j,
    \qquad
    \tau_{\mathrm{ref}}^2
    =
    \frac{1}{W}
    \sum_{j=1}^{W}
    \tau_j^2 .
\end{equation}

The warmup statistics provide an early-training reference for
reliability estimation. Instead of using an absolute signal threshold,
SAGE measures the relative change of the current reliability with
respect to this reference, allowing the shrinkage factor to adapt to
the temporal variation of the privatized estimates.

\begin{figure*}[t]
    \centering
    \includegraphics[width=\textwidth]
    {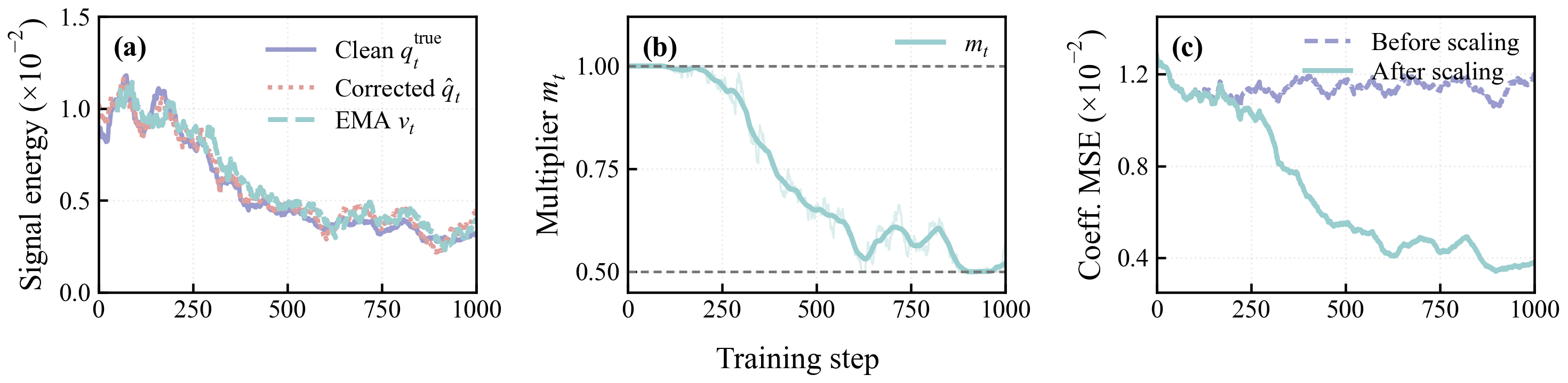}
    \caption{
    Diagnostics of SAGE on a representative RoBERTa-large SNLI run
    under \(\epsilon=6\) and \(K=64\).
    (a) The corrected statistic \(\hat q_t\) and its EMA \(v_t\)
    track the clean aggregate energy
    \(q_t^{\mathrm{true}}=\|u_t\|_2^2/K\).
    (b) The multiplier \(m_t\) adapts within \([0.5,1]\).
    (c) Adaptive scaling reduces the error to the clean aggregate by
    \(42.3\%\) after warmup and \(65.5\%\) over the final 200 steps.
    The clean aggregate \(u_t\) is used only for offline diagnostics
    and is never accessed by SAGE.
    }
    \label{fig:sage_diagnostics}
\end{figure*}

Based on the tracked signal energy, SAGE constructs a bounded
reliability score:
\begin{equation}
    \phi_t
    =
    \frac{v_t}
    {v_t+\lambda\tau_t^2},
\end{equation}
where \(\lambda>0\) controls the contribution of the known
noise level.

The reliability score follows an increasing-saturating form: it
increases with the estimated signal energy and gradually approaches one
when the released estimate becomes sufficiently reliable. However,
directly using \(\phi_t\) as the scaling factor would introduce a
global dependence on the absolute magnitude of the reliability
estimate, which varies across tasks, privacy settings, and optimization
stages. SAGE therefore adopts a relative reliability measure with
respect to the early-training reference:
\begin{equation}
    \phi_{\mathrm{ref}}
    =
    \frac{v_{\mathrm{ref}}}
    {v_{\mathrm{ref}}+\lambda\tau_{\mathrm{ref}}^2}.
\end{equation}

The final scaling multiplier is obtained by comparing the current
reliability with this reference:
\begin{equation}
    m_t
    =
    \mathrm{clip}
    \left(
    \frac{\phi_t}{\phi_{\mathrm{ref}}},
    m_{\min},
    1
    \right).
\end{equation}
The lower bound prevents excessive attenuation, while the
upper bound ensures that SAGE never amplifies the released
estimates. The released privatized estimate vector is then
scaled as
\begin{equation}
    \bar{s}_t
    =
    m_t\tilde{s}_t .
\end{equation}

Since \(m_t\) depends only on privatized outputs and mechanism parameters, SAGE is pure post-processing and
requires neither additional privacy budget nor forward
queries.

\subsection{Optimization Interpretation of Reliability-Aware Scaling}
\label{sec:shrinkage_effect}

We analyze why attenuating unreliable privatized estimates
can improve optimization. Consider the zeroth-order update
after applying a scalar multiplier:
\begin{equation}
    \Delta\theta_t
    =
    -\eta m_t(G_t+N_t),
\end{equation}
where \(G_t\) denotes the clean random-direction ZO
estimator and \(N_t\) denotes the update perturbation induced
by Gaussian privacy noise after aggregation over random
directions.

Under the standard smoothness assumption and
ignoring higher-order finite-difference errors, the expected
one-step loss change satisfies
\begin{equation}
    \mathbb{E}[\Delta\mathcal L_t]
    \lesssim
    -\eta m_t S_t
    +
    \frac{L\eta^2m_t^2}{2}
    \left(R_KS_t+N_K\right),
\end{equation}
where \(S_t=\|g_t\|_2^2\) denotes the underlying
optimization-signal magnitude, \(R_K\) characterizes the
variance introduced by random-direction estimation, and
\(N_K\) represents the contribution of Gaussian privacy noise
after random-direction aggregation.

The first term represents the expected descent contributed by
the useful optimization signal, whereas the second term
captures the update risk induced by random-direction
estimation and privacy noise. Importantly, the two terms
respond differently to scaling: the descent term decreases
linearly with \(m_t\), while the update-risk term decreases
quadratically with \(m_t\).

Ignoring the constraint \(m_t\leq1\), minimizing the
one-step upper bound with respect to \(m_t\) gives the oracle
scaling magnitude
\begin{equation}
    m_t^\star
    =
    \frac{S_t}
    {L\eta\left(R_KS_t+N_K\right)}.
\end{equation}
This oracle quantity is unavailable because the clean signal
and privacy perturbation are not separately observable, but
it indicates that the preferred scale should increase with
estimate reliability.

SAGE follows this principle using the computable score
\begin{equation}
    \phi_t
    =
    \frac{v_t/\tau_t^2}
    {v_t/\tau_t^2+\lambda},
\end{equation}
which is increasing and saturating in the estimated
signal-to-noise level. The warmup-referenced multiplier
therefore applies conservative attenuation only after the
estimated reliability falls below its reference value.

Figure~\ref{fig:sage_diagnostics} examines estimator calibration,
controller adaptation, and the effect of scaling. In panel~(a), the
corrected statistic closely tracks the temporal variation of the clean
aggregate energy. For the displayed 50-step trajectories, its
normalized mean absolute error and normalized root mean squared error
are \(7.3\%\) and \(7.9\%\), respectively, with a correlation of
\(0.987\).

As the estimated reliability decreases, panel~(b) shows that SAGE
gradually reduces \(m_t\) rather than applying a fixed multiplier.
Panel~(c) compares
\begin{equation}
    \frac{1}{K}\|\tilde{s}_t-u_t\|_2^2
    \quad\text{and}\quad
    \frac{1}{K}\|m_t\tilde{s}_t-u_t\|_2^2,
\end{equation}
showing that adaptive scaling reduces the post-warmup error by
\(42.3\%\), with a \(65.5\%\) reduction over the final 200 steps.
Together, these diagnostics show that SAGE tracks reliability changes
and improves the fidelity of the privatized aggregate in this
representative run.

\section{Experiments}
\label{sec:experiments}

\begin{table*}[!t]
\centering

{\setlength{\tabcolsep}{5pt}
\begin{tabular}{llccccccc}
\toprule
\multirow{2}{*}{\textbf{Privacy constraint}}
& \multirow{2}{*}{\textbf{Algorithm}}
& \multicolumn{2}{c}{\textbf{Sentiment}}
& \multicolumn{3}{c}{\textbf{Natural Language Inference}}
& \multicolumn{1}{c}{\textbf{Topic}}
& \multicolumn{1}{c}{\textbf{Avg.}} \\
\cmidrule(lr){3-4}
\cmidrule(lr){5-7}
\cmidrule(lr){8-8}
\cmidrule(lr){9-9}
& & \textbf{SST-2} & \textbf{SST-5} & \textbf{SNLI} & \textbf{MNLI}
& \textbf{RTE} & \textbf{TREC} & \textbf{Avg.} \\
\midrule
\multirow{4}{*}{$\epsilon=2, \delta=10^{-5}$}
& DP-AdamW & 91.3 & 47.4 & 74.3 & 73.6 & 72.5 & 91.1 & 75.03 \\
& DPZero & 92.3 & 46.6 & 73.4 & 62.2 & 70.2 & 83.3 & 71.33 \\
& DP-AggZO ($K=64$) & 92.2 & 50.6 & 77.5 & 73.7 & 72.2 & 92.1 & 76.38 \\
& SAGE ($K=64$) & \textbf{93.2} & \textbf{53.0} & \textbf{78.7}
& \textbf{75.3} & \textbf{75.1} & \textbf{94.2} & \textbf{78.25} \\
\midrule
\multirow{4}{*}{$\epsilon=6, \delta=10^{-5}$}
& DP-AdamW & 92.3 & 48.8 & 81.4 & 76.4 & 77.1 & 92.3 & 78.05 \\
& DPZero & 91.8 & 48.9 & 77.4 & 67.5 & 71.5 & 87.2 & 74.05 \\
& DP-AggZO ($K=64$) & 92.7 & 50.8 & 82.6 & 77.1 & 77.2 & 94.0 & 79.07 \\
& SAGE ($K=64$) & \textbf{93.4} & \textbf{54.4} & \textbf{85.1}
& \textbf{78.4} & \textbf{78.8} & \textbf{94.6} & \textbf{80.78} \\
\midrule
\multirow{1}{*}{Perfect privacy}
& Zero-Shot
& 79.0
& 35.5
& 50.2
& 48.8
& 51.4
& 32.0
& 49.48
\\
\bottomrule
\end{tabular}}
\caption{Test performance on RoBERTa (355M). The best result is highlighted in bold.}
\label{tab:main_results}
\end{table*}

\begin{table*}[!t]
\centering

{\setlength{\tabcolsep}{6pt}
\begin{tabular}{llcccc}
\toprule
\multirow{2}{*}{\textbf{Privacy constraint}}
& \multirow{2}{*}{\textbf{Algorithm}}
& \multicolumn{2}{c}{\textbf{SST-2}}
& \multicolumn{2}{c}{\textbf{SQuAD}} \\
\cmidrule(lr){3-4}
\cmidrule(lr){5-6}
& & \textbf{OPT-1.3B} & \textbf{OPT-6.7B}
& \textbf{OPT-1.3B} & \textbf{OPT-6.7B} \\
\midrule
\multirow{6}{*}{$\epsilon=2, \delta=10^{-5}$}
& DP-AdamW & 91.2 & OOM & 77.7 & OOM \\
& DPZero & 87.1 & 92.2 & 72.0 & 78.1 \\
& DP-AggZO ($K=16$) & 90.4 & 93.4 & 76.8 & 83.0 \\
& DP-AggZO ($K=64$) & 90.9 & 94.1 & 78.8 & 83.2 \\
& SAGE ($K=16$) & 91.1 & 94.1 & 77.9 & 83.9 \\
& SAGE ($K=64$) & \textbf{91.5} & \textbf{94.4} & \textbf{80.1} & \textbf{84.6} \\
\midrule
\multirow{6}{*}{$\epsilon=6, \delta=10^{-5}$}
& DP-AdamW & 91.3 & OOM & 79.5 & OOM \\
& DPZero & 87.9 & 92.8 & 74.0 & 80.0 \\
& DP-AggZO ($K=16$) & 91.1 & 94.5 & 77.2 & 83.1 \\
& DP-AggZO ($K=64$) & 91.2 & \textbf{94.8} & 79.3 & 84.2 \\
& SAGE ($K=16$) & 91.6 & 94.4 & 78.5 & 84.3 \\
& SAGE ($K=64$) & \textbf{92.1} & 94.6 & \textbf{80.8} & \textbf{85.6} \\
\midrule
Perfect privacy & Zero-Shot & 53.6 & 61.2 & 26.8 & 36.5 \\
\bottomrule
\end{tabular}}
\caption{Test performance on OPT-1.3B and OPT-6.7B. The best result is highlighted in bold. OOM stands for out of memory.}
\label{tab:opt_results_scaleema}
\end{table*}

We evaluate SAGE for differentially private full-parameter
fine-tuning of RoBERTa-large, OPT-1.3B, and OPT-6.7B
across text classification and question-answering tasks.
We compare SAGE with representative first-order and
zeroth-order private fine-tuning methods, using DP-AggZO
as the primary baseline. For matched comparisons at a fixed
number of directions \(K\), DP-AggZO and SAGE use the
same privacy budget, number of forward evaluations, training
steps, and hyperparameter-tuning budget. Since SAGE
operates entirely as post-processing of the privatized
coefficient vector, it introduces no additional private queries
or privacy cost. We further analyze the temporal behavior of
aggregate reliability and characterize the computation and
state overhead introduced by SAGE. All experiments are
conducted on NVIDIA GeForce RTX 5090 GPUs.

\subsection{Experimental Setup}
\label{sec:experimental_setup}

\paragraph{Models and tasks.}
Following DP-AggZO,
we evaluate full-parameter private fine-tuning on
RoBERTa-large (355M)~\citep{liu2019robertarobustlyoptimizedbert},
OPT-1.3B, and OPT-6.7B~\citep{zhang2022optopenpretrainedtransformer}.
RoBERTa-large is evaluated on SST-2, SST-5~\citep{socher-etal-2013-recursive}, SNLI~\citep{bowman-etal-2015-large},
MNLI~\citep{williams-etal-2018-broad}, RTE~\citep{wang-etal-2018-glue}, and TREC~\citep{li-roth-2002-learning}, while OPT models are evaluated
on SST-2 and SQuAD~\citep{rajpurkar-etal-2016-squad}.
We follow the same data splits and evaluation protocols
as DP-AggZO, reporting accuracy for classification tasks
and F1 score for SQuAD.

\paragraph{Baselines.}
We compare SAGE with DPZero, DP-AdamW, and DP-AggZO. DP-AggZO is the primary baseline since SAGE only performs post-processing on its privatized coefficient vector. Zero-shot evaluation is included as non-private references. All private results in Tables~\ref{tab:main_results} and
\ref{tab:opt_results_scaleema} are obtained using our codebase.
For matched comparisons, DP-AggZO and SAGE use the same privacy
budget, query budget, number of training steps, and value of \(K\).
Their method-specific hyperparameter search spaces are reported in
Supp. 3.

\paragraph{Training configurations.}
We consider $(\epsilon,\delta)$-DP with
$\delta=10^{-5}$ and $\epsilon\in\{2,6\}$.
Privacy accounting follows DP-AggZO using the
Opacus RDP accountant.
Unless otherwise specified, experiments use
$T=1000$ updates, target batch size 64,
and perturbation scale $\mu=10^{-3}$.
We use $K=64$ directions for RoBERTa-large and
$K\in\{16,64\}$ for OPT models.

SAGE uses fixed controller parameters
$\beta=0.03$, $W=50$, $\lambda=1$,
$\rho=10^{-8}$, and $m_{\min}=0.5$
across all tasks. 

\subsection{Main Results}
\label{sec:main_results}

\paragraph{SAGE improves private fine-tuning on
RoBERTa-large.}
Table~\ref{tab:main_results} reports test accuracy on six
classification benchmarks. SAGE achieves the highest average
accuracy among all private methods under both privacy budgets,
outperforming the first-order DP-AdamW, the single-direction
DPZero, and the aggregation-based DP-AggZO.

Under \(\epsilon=6\), SAGE obtains an average accuracy of
\(80.78\%\), exceeding DP-AdamW, DPZero, and DP-AggZO by
\(2.73\), \(6.73\), and \(1.72\) percentage points,
respectively. In the matched comparison with DP-AggZO,
SAGE improves all six tasks, with the largest gains on SST-5
and SNLI (\(+3.6\) and \(+2.5\) points).
Under the stronger privacy constraint \(\epsilon=2\), SAGE
again achieves the best private average of \(78.25\%\),
improving DP-AggZO by \(1.87\) points and outperforming it
on all six benchmarks. The largest gains are observed on RTE,
SST-5, and TREC, with improvements of \(2.9\), \(2.4\), and
\(2.1\) percentage points, respectively. These improvements
require neither additional private queries nor changes to the
underlying privacy mechanism.

\paragraph{SAGE scales to larger language models.}
Table~\ref{tab:opt_results_scaleema} evaluates SAGE on OPT-1.3B and OPT-6.7B for SST-2 and SQuAD. SAGE consistently outperforms DPZero across model sizes, tasks, and privacy budgets,
achieving the best or near-best private performance. With \(K=64\) and \(\epsilon=2\),
SAGE improves DP-AggZO on all four model--task combinations, with gains of \(0.6\) and \(0.3\) points on SST-2 and \(1.3\) and \(1.4\) points on SQuAD for OPT-1.3B and OPT-6.7B, respectively.

Under \(\epsilon=6\), SAGE improves three of the four
matched \(K=64\) settings, including gains of \(1.5\) and
\(1.4\) F1 points on SQuAD, while remaining within
\(0.2\) points of DP-AggZO on OPT-6.7B SST-2.
It also surpasses DP-AdamW in all settings where the latter
is executable, whereas DP-AdamW runs out of memory on
OPT-6.7B. These results show that SAGE preserves the
scalability of forward-only DP-ZO while improving the utility
of the same privatized estimates.

\subsection{Ablation Study}
\label{sec:ablation}

We conduct ablation studies on RoBERTa-large to
investigate the contribution of each component in SAGE.
All experiments are evaluated under
\(\epsilon=6\) with \(K=64\).
We consider three key components:
(i) noise-floor correction, which removes the known
Gaussian variance from the released aggregate energy;
(ii) EMA tracking, which stabilizes the reliability estimate
over training iterations; and
(iii) warmup-based relative calibration, which provides
a task-dependent reliability reference for shrinkage. 

\paragraph{Effect of noise correction and temporal tracking.}
Table~\ref{tab:ablation_components}
shows the effect of removing individual components.
Removing noise-floor correction consistently degrades
performance, because the reliability estimator becomes
biased by the additive privacy noise and tends to
overestimate the signal strength.
Removing EMA tracking also decreases utility, indicating
that instantaneous reliability estimates contain substantial
iteration-level fluctuations.
The complete SAGE achieves the best average performance,
demonstrating that accurate and stable reliability estimation
is important for effective shrinkage.

\begin{table}[t]
\centering

\begin{tabular}{lcccc}
\toprule
\textbf{Method}
& \textbf{SST-5}
& \textbf{SNLI}
& \textbf{MNLI}
& \textbf{Avg.}
\\
\midrule

SAGE
& \textbf{54.4}
& \textbf{85.1}
& \textbf{78.4}
& \textbf{72.6}
\\

w/o noise correction
& 53.5
& 84.3
& 77.3
& 71.7
\\

w/o EMA tracking
& 53.9
& 84.6
& 77.5
& 72.0
\\

w/o warmup anchor
& 53.6
& 84.2
& 77.0
& 71.6
\\

\bottomrule
\end{tabular}
\caption{
Ablation study of noise correction and temporal tracking
in SAGE on RoBERTa-large.
Results are evaluated under $\epsilon=6$ with $K=64$.
}
\label{tab:ablation_components}
\end{table}

\paragraph{Effect of warmup reference calibration.}

\begin{table}[t]
\centering

\begin{tabular}{lcccc}
\toprule
\textbf{Warmup steps}
& \textbf{SST-5}
& \textbf{SNLI}
& \textbf{MNLI}
& \textbf{Avg.}
\\
\midrule

$W=25$
& 54.0
& 84.7
& 77.8
& 72.2
\\

$W=50$ (default)
& \textbf{54.4}
& \textbf{85.1}
& \textbf{78.4}
& \textbf{72.6}
\\

$W=75$
& 54.3
& 84.9
& 78.1
& 72.4
\\

$W=100$
& 54.1
& 84.8
& 77.9
& 72.3
\\

\bottomrule
\end{tabular}
\caption{
Effect of warmup window in SAGE on RoBERTa-large.
Results are evaluated under $\epsilon=6$ with $K=64$.
}
\label{tab:warmup_sensitivity}
\end{table}

Table~\ref{tab:warmup_sensitivity}
studies the influence of the warmup window.
Using a shorter warmup period provides a less stable
reference due to insufficient statistics, whereas an overly
long warmup delays the adaptation of the shrinkage controller.
The default \(W=50\) achieves the best overall performance,
showing that a moderate warmup period provides a reliable
balance between reference stability and adaptation speed.

\subsection{Computation and Memory Efficiency}
\label{sec:efficiency}

\paragraph{SAGE preserves the query and state efficiency
of DP-AggZO.}
DP-AggZO and SAGE require exactly the same \(2K\)
forward evaluations per parameter update. SAGE adds only
coefficient-space post-processing: computing
\(\|\tilde{s}_t\|_2^2\), subtracting the known Gaussian
noise floor, updating a scalar EMA state, evaluating the
shrinkage multiplier, and rescaling the \(K\)-dimensional
coefficient vector. These additional operations require only
\(O(K)\) arithmetic per step and no model evaluations.

The persistent state introduced by SAGE consists of a
constant number of scalar statistics, including \(v_t\),
\(v_{\mathrm{ref}}\), \(\tau_{\mathrm{ref}}^2\), and
\(\phi_{\mathrm{ref}}\). It therefore requires \(O(1)\)
additional optimizer state and introduces no
parameter-dimensional gradient, momentum, or
second-moment buffers. SAGE consequently preserves the
forward-only and memory-efficient structure of DP-AggZO
while providing reliability-aware post-processing of its
privatized coefficient aggregate. Table~\ref{tab:memory} confirms no extra memory cost for SAGE.

\begin{table}[t]
\centering

\setlength{\tabcolsep}{6pt}
\begin{tabular}{cccc}
\toprule
\textbf{Model}
& \textbf{Dataset}
& \textbf{Algorithm}
& \textbf{Memory}
\\
\midrule

\multirow{3}{*}{RoBERTa}
& \multirow{3}{*}{SST-2}
& DP-AdamW
& \textbf{11.38 GiB}
\\
&
&
DP-AggZO
& 2.99 GiB
\\
&
&
SAGE
& 2.99 GiB
\\

\midrule

\multirow{3}{*}{RoBERTa}
& \multirow{3}{*}{MNLI}
& DP-AdamW
& \textbf{12.35 GiB}
\\
&
&
DP-AggZO
& 3.25 GiB
\\
&
&
SAGE
& 3.25 GiB
\\

\midrule

\multirow{3}{*}{RoBERTa}
& \multirow{3}{*}{SNLI}
& DP-AdamW
& \textbf{11.39 GiB}
\\
&
&
DP-AggZO
& 2.99 GiB
\\
&
&
SAGE
& 2.99 GiB
\\

\bottomrule
\end{tabular}
\caption{
Peak memory usage of DP-AdamW, DP-AggZO and SAGE, tested on a 5090 GPU.
The memory usage of DP-AggZO with different $K$ values is identical.
}
\label{tab:memory}
\end{table}

\section{Conclusion}
\label{sec:conclusion}

In this work, we propose SAGE, a noise-aware shrinkage method for differentially private zeroth-order fine-tuning of large language models. SAGE estimates the time-varying signal quality of privatized zeroth-order coefficients by correcting their observed energy for the known Gaussian noise floor, and then adaptively scales their contribution through bounded post-processing. Our analysis shows that such shrinkage reduces the quadratic update-risk term more strongly than the useful descent term, providing a principled explanation for attenuating noise-dominated updates. Experiments across multiple language models and downstream tasks demonstrate that SAGE consistently improves the utility of DP-ZO fine-tuning under matched privacy budgets, query budgets, and training settings, while preserving its forward-only and memory-efficient structure.

\bibliography{references}

\end{document}